\documentclass[conference]{IEEEtran}
\IEEEoverridecommandlockouts
\usepackage{cite}
\usepackage{amsmath,amssymb,amsfonts}

\usepackage{graphicx}
\usepackage{textcomp}
\usepackage{xcolor}
\usepackage{algorithm}
\usepackage{algpseudocode}
\usepackage{tabularx}
\usepackage{booktabs}
\usepackage{array}
\usepackage{xcolor}
\usepackage{graphicx}
\usepackage{subcaption}
\usepackage{algorithm} 
\usepackage{algpseudocode}
\definecolor{forestgreen}{RGB}{34,139,34}
\usepackage{enumitem}

\usepackage[utf8]{inputenc}   
\usepackage[T1]{fontenc}
\usepackage{textcomp} 
\def\BibTeX{{\rm B\kern-.05em{\sc i\kern-.025em b}\kern-.08em
    T\kern-.1667em\lower.7ex\hbox{E}\kern-.125emX}}
\begin{document}

\title{Proof2Silicon: Prompt Repair for Verified Code and Hardware Generation via Reinforcement Learning \\
\thanks{This work was presented at SRC TECHCON 2025.}
 }

\author{\IEEEauthorblockN{Manvi Jha}
\IEEEauthorblockA{\textit{Electrical and Computer Engineering} \\
\textit{University of Illinois Urbana-Champaign}\\
Urbana, Illinois, USA \\
manvij2@illinois.edu}
\and
\IEEEauthorblockN{Jiaxin Wan}
\IEEEauthorblockA{\textit{Electrical and Computer Engineering} \\
\textit{University of Illinois Urbana-Champaign}\\
Urbana, Illinois, USA \\
wan25@illinois.edu}
\and
\IEEEauthorblockN{Deming Chen}
\IEEEauthorblockA{\textit{Electrical and Computer Engineering} \\
\textit{University of Illinois Urbana-Champaign}\\
Urbana, Illinois, USA \\
dchen@illinois.edu}
}

\maketitle


\begin{abstract}
Large Language Models (LLMs) have demonstrated impressive capabilities in automated code generation but frequently produce code that fails formal verification, an essential requirement for hardware and safety-critical domains. To overcome this fundamental limitation, we previously proposed PREFACE, a model-agnostic framework based on reinforcement learning (RL) that iteratively repairs the prompts provided to frozen LLMs, systematically steering them toward generating formally verifiable Dafny code without costly fine-tuning. This work presents Proof2Silicon, a novel end-to-end synthesis framework that embeds the previously proposed PREFACE flow to enable the generation of correctness-by-construction hardware directly from natural language specifications. Proof2Silicon operates by: (1) leveraging PREFACE’s verifier-driven RL agent to optimize prompt generation iteratively, ensuring Dafny code correctness; (2) automatically translating verified Dafny programs into synthesizable high-level C using Dafny’s Python backend and PyLog; and (3) employing Vivado HLS to produce RTL implementations. Evaluated rigorously on a challenging 100-task benchmark, PREFACE's RL-guided prompt optimization consistently improved Dafny verification success rates across diverse LLMs by up to 21\%. Crucially, Proof2Silicon achieved an end-to-end hardware synthesis success rate of up to 72\%, generating RTL designs through Vivado HLS synthesis flows. These results demonstrate a robust, scalable, and automated pipeline for LLM-driven, formally verified hardware synthesis, bridging natural-language specification and silicon realization.
\end{abstract}


\section{Introduction}

Large Language Models (LLMs) have significantly advanced automated code generation from natural language prompts \cite{10.1145/3643795.3648384}. Although large language models can sometimes produce syntactically valid code, they still fall short of guaranteeing complete correctness. This shortcoming affects every application, but it is particularly unacceptable in contexts where undetected errors can have catastrophic consequences (for example, cryptographic libraries, aerospace control software, automotive safety systems, and hardware design \cite{10617643} \cite{murphy2024combiningllmcodegeneration}). In such domains, formal methods and provable correctness are essential to ensure reliability and security. When LLMs are tasked with generating code in formally verified languages such as Dafny \cite{10.1007/978-3-642-17511-4_20}, even small mismatches in the specified contracts or the omission of critical invariants can cause the output to fail verification, or worse, introduce undetected unsoundness.

Traditional solutions to enhance LLM performance in such specialized contexts involve fine-tuning \cite{ren2025learningdynamicsllmfinetuning}, which adapts models to domain-specific tasks. However, fine-tuning large models requires substantial computational resources, extensive annotated data, and risks overwriting beneficial pre-trained capabilities, leading to brittleness and inefficiency.

In contrast, we argue that the latent reasoning abilities inherent within LLMs are sufficient to address many domain-specific tasks if effectively guided through prompts. The quality of the prompts significantly impacts the effectiveness of LLM in formal verification tasks. Constructing these prompts manually is arduous, particularly when feedback from verification engines needs to be integrated into subsequent attempts. However, this structured feedback from formal verifiers offers precisely the actionable insights needed to refine model behavior more systematically.

Motivated by this insight, we introduced PREFACE \cite{10.1145/3716368.3735300}, our lightweight and scalable RL framework in which a Small Language Model (SLM) is trained to generate optimized prompts for a fixed, general-purpose LLM, allowing accurate, formally verified code synthesis without costly fine-tuning of the base model. Building on PREFACE, we then present Proof2Silicon, which embeds this same RL-driven prompt adaptation mechanism into a full‑stack high‑level synthesis pipeline: it takes the verified code output from the LLM and automatically transforms it into synthesizable HLS C (via Dafny’s Python backend and PyLog \cite{9591456}) and ultimately into RTL through Vivado HLS. The details of PREFACE and its integration into the Proof2Silicon workflow are described in Sections III and IV.

Focusing on Dafny, a verification-aware programming language that facilitates formal specifications through built-in constructs such as preconditions, postconditions, invariants, and lemmas, we bridge verified software generation with hardware design workflows. Dafny automatically translates formal specifications into verification conditions checked by SMT solvers, thereby enabling correctness-by-construction software development. Given the scarcity of domain-specific datasets (for example, DafnyBench \cite{loughridge2024dafnybenchbenchmarkformalsoftware} has only \textasciitilde700 tasks), conventional fine-tuning is impractical. Proof2Silicon addresses this challenge by taking advantage of the existing capabilities of a frozen LLM and steering it through an RL-driven prompt refinement loop (via the embedded PREFACE flow), allowing efficient verified code generation. 
Verified Dafny programs are systematically translated into HLS code, bridging formal methods directly into hardware synthesis workflows, building on decades of HLS advancements \cite{xpilot}.

\subsection{Contributions}

This work presents a novel reinforcement learning-based framework for guiding LLMs toward generating formally verified code without requiring fine-tuning. Our key contributions are as follows.
\begin{itemize}[%
  leftmargin=0.8em,       
  label=\textbullet,    
  itemindent=0.2em      
]
    \item \textbf{End-to-end verified hardware code generation:} We propose a full-stack pipeline—from natural language specifications to verified Dafny code, systematically translated into HLS C code, ensuring correctness-by-construction for hardware implementation without the need for costly fine-tuning or a dedicated dataset.
    \item \textbf{Verifier-guided prompt refinement via RL:} A small, RL-trained agent iteratively optimizes prompts based on formal verification feedback, significantly enhancing verification success without modifying the underlying LLM. This utilizes the PREFACE model previously proposed by us.
    \item \textbf{A formal verification feedback loop:} We design a generation verification loop that uses structured feedback from an integrated Dafny verifier. The RL agent interprets the error metadata and learns to iteratively refine the commands to improve the success of the verification.
    \item \textbf{Formal verification integrated into hardware synthesis:} Verified Dafny code is converted through a Dafny → Python → Customized Python → HLS C pipeline, validated with Vivado HLS, thus closing the gap between software correctness and hardware design.
    \item \textbf{Significant empirical improvements and synthesis success:} Our framework achieves up to a 21\% increase in verification success rates across diverse LLMs on a benchmark of 100 tasks, with upto 72\% of verified programs successfully synthesized into RTL through Vivado HLS.

\end{itemize}

Collectively, these contributions establish Proof2Silicon as a scalable and efficient alternative to traditional fine-tuning approaches, enabling robust, formally verified hardware generation directly from natural language specifications.

\section{Background and Related Work}

\subsection{Dafny: Syntax and Verification Workflow}

Dafny \cite{10.1007/978-3-642-17511-4_20} is a verification-aware imperative programming language integrating formal specifications directly into its syntax. Methods define explicit contracts through preconditions (\textit{requires}) and postconditions (\textit{ensures}), establishing strict adherence to specifications. Loop constructs enforce properties using invariants, which must hold true throughout loop iterations, providing rigorous runtime correctness guarantees. Dafny utilizes \textit{ghost variables} and \textit{lemma} functions to support purely logical reasoning constructs, enabling verification without incurring runtime overhead. Internally, Dafny compiles these specifications into first-order verification conditions, subsequently discharged automatically by Satisfiability Modulo Theories (SMT) solvers such as Z3 \cite{10.1007/978-3-662-46081-8_5}. This ensures correctness-by-construction, making Dafny particularly suitable for high-assurance, LLM-driven hardware synthesis tasks.

\subsection{Dafny Generation \& Verification with LLMs}

Recent studies have explored LLM capabilities in generating formally verified Dafny programs \cite{NEURIPS2023_43e9d647,10617643,ma2025specgenautomatedgenerationformal}. Misu et al. \cite{10.1145/3643763} leveraged structured chain-of-thought prompting with GPT-4, enhancing Dafny program verification rates by methodically guiding the model through logical deduction paths, significantly outperforming naive prompting methods that lack structured reasoning.

Poesia et al. \cite{poesia2024dafnyannotatoraiassistedverificationdafny} proposed Dafny-Annotator, an iterative annotation refinement framework combining LLM predictions with greedy search methods to determine logical annotations. To tackle data scarcity, they further introduced DafnySynth, an automated dataset generation pipeline that produces synthetic Dafny tasks to augment training data, enabling fine-tuned models to achieve substantial improvements on benchmarks like DafnyBench \cite{loughridge2024dafnybenchbenchmarkformalsoftware}.

Li et al. \cite{li2025dafnyverificationawareintermediatelanguage} presented an end-to-end pipeline treating Dafny as a verification-aware intermediate language. Their framework generates Dafny programs directly from natural language prompts, facilitating iterative refinement through formal verification feedback loops. Verified Dafny code is then automatically compiled into Python, significantly simplifying specification-coding interactions and improving user accessibility.

Proof2Silicon builds on these developments by embedding PREFACE, which introduces reinforcement learning (RL)-driven prompt optimization. Differing from direct fine-tuning approaches, it trains an RL-based small language model (SLM) to iteratively refine prompts, effectively steering frozen, general-purpose LLMs toward generating formally verified Dafny code. Embedding this novel integration of reinforcement learning and formal verification frameworks with our Dafny to RTL workflow, distinctly positions Proof2Silicon as a robust, scalable solution to generating formally verified Dafny code for hardware design.

\begin{figure*}[t]
    \centering
    \includegraphics[width=1.01\linewidth]{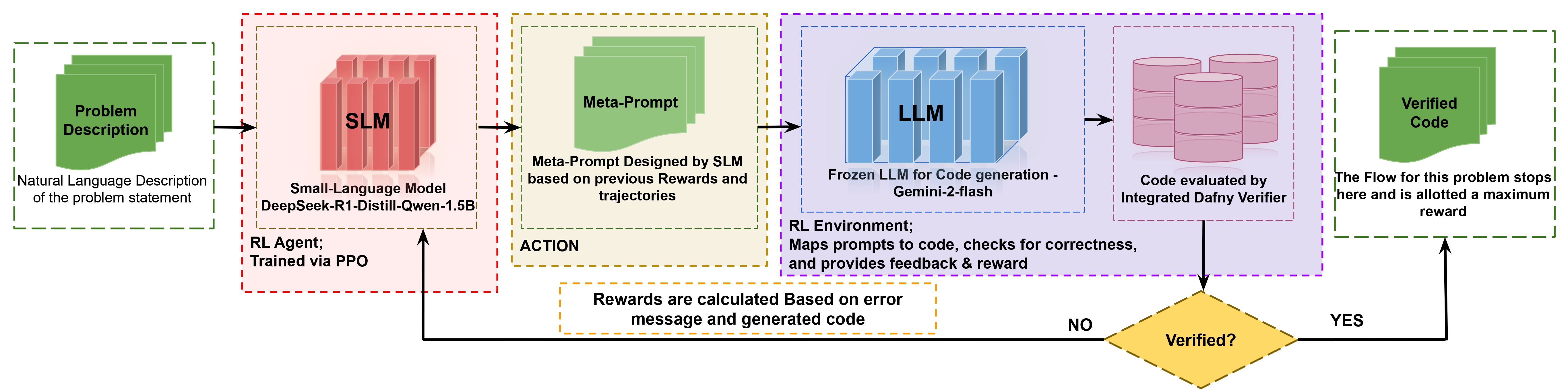}
    \caption{Overview of the PREFACE \cite{10.1145/3716368.3735300} framework-forming the prompt‑optimization core embedded within the Proof2Silicon pipeline.}
    \label{fig:rl_framework}
\end{figure*}

\subsection{LLMs for RL and Prompt Optimization}

Recent research incorporates LLMs into reinforcement learning workflows, manifesting primarily as either high-level policy guidance or direct RL-driven prompt optimization.

Du et al. \cite{du2023guidingpretrainingreinforcementlearning} leveraged LLMs as high-level reasoning modules within RL systems, guiding RL agents through sub-goal generation and exploration strategies to accelerate and enhance learning outcomes. In multi-agent scenarios, frameworks like YOLO-MARL \cite{zhuang2024yolomarlllmmultiagentreinforcement} and LERO \cite{wei2025lerollmdrivenevolutionaryframework} utilize LLMs to produce strategic policy sketches or dynamic reward functions, significantly improving collaborative task performance.

In contrast, Proof2Silicon focuses specifically on RL-driven prompt optimization. It formulates prompt generation as an iterative RL problem, employing verifier feedback as a direct reward signal. By continuously optimizing prompts, Proof2Silicon efficiently leverages the latent reasoning capabilities of frozen LLMs, thereby achieving high verification rates without expensive retraining or fine-tuning.

\subsection{LLM-assisted Hardware Code Generation}

Integrating LLMs into hardware design automation, especially for Register Transfer Level (RTL) and High-Level Synthesis (HLS), has attracted significant research attention. Early works primarily targeted direct RTL generation from natural language prompts or structured specifications \cite{liu2024rtlcoderfullyopensourceefficient,yu2025spec2rtlagentautomatedhardwarecode}. However, direct RTL generation methods frequently encountered semantic ambiguities, limiting practical scalability and correctness \cite{10.1145/3676536.3676781}.

Advanced frameworks, such as Spec2RTL-Agent \cite{yu2025spec2rtlagentautomatedhardwarecode} and HLSPilot \cite{10.1145/3676536.3676781}, proposed sophisticated multi-stage generation pipelines. These approaches first translate natural language descriptions into structured, synthesizable intermediate languages (e.g., synthesizable C++), then utilize conventional HLS tools to ensure accurate hardware realization. RTLCoder \cite{liu2024rtlcoderfullyopensourceefficient} further enhanced these pipelines by introducing automated dataset generation techniques specifically designed for RTL synthesis tasks, significantly surpassing general-purpose models such as GPT-3.5.

VerilogReader \cite{10691801} pioneered an advanced, coverage-directed hardware test generation strategy, utilizing LLMs to parse RTL descriptions precisely and enhance testbench coverage dramatically compared to traditional random testing. FormalEval \cite{10617643} introduced automated formal verification using sequential equivalence checking, offering accurate, fully automated verification for RTL code generated by LLMs, significantly enhancing verification robustness.

CreativEval \cite{10691798} provided a novel approach to assess the creativity of LLM-generated hardware designs. It established comprehensive cognitive evaluation metrics—including fluency, flexibility, originality, and elaboration—beyond traditional correctness measures, highlighting LLM potential for innovative and robust hardware design.

Proof2Silicon uniquely integrates these advancements, employing an RL-driven prompt optimization pipeline tightly coupled with formal verification and HLS tools. This comprehensive integration facilitates the direct translation from natural language to verified Dafny code, and ultimately to synthesizable hardware descriptions, positioning Proof2Silicon as a state-of-the-art methodology for reliable, formally verified hardware synthesis.

\begin{figure*}[t]
    \centering
    \includegraphics[width=.75\linewidth]{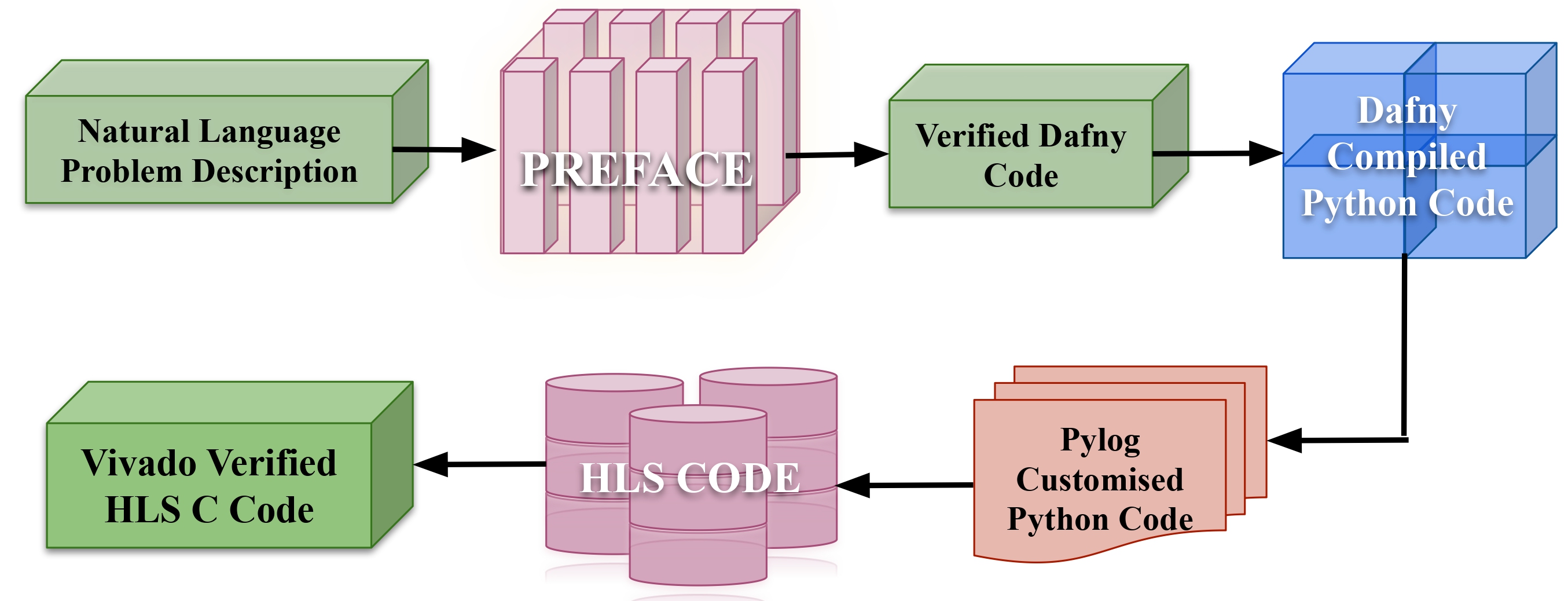}
    \caption{Overview of the proposed Proof2Silicon framework.}
    \label{fig:proposed_framework}
\end{figure*}

\section{PREFACE Overview}

\subsection{Pipeline Architecture}
PREFACE \cite{10.1145/3716368.3735300} is an RL-guided pipeline developed to synthesize formally verified Dafny code, a language designed for rigorous specification and automated verification. It centers on an RL‑driven Small Language Model (SLM) that refines the prompts to a frozen LLM \(\mathcal{M}\) for the synthesis of Dafny code. Given task specifications \(\tau\) (pre / postconditions), a handcrafted prompt \(p_{0}=\varphi(\tau)\) is submitted to \(\mathcal{M}\), producing a candidate code \(c_{t}\). The Dafny verifier then computes the error count \(e_{t}\) and feedback \(o_{t}\). On success (\(e_{t}=0\)), the loop terminates; otherwise, the SLM ingests \((p_{t},c_{t},o_{t})\) to sample a prompt edit \(\Delta p_{t}\), yielding \(p_{t+1}=p_{t}\oplus\Delta p_{t}\). This cycle repeats for up to \(T_{\max}\) iterations.

\begin{algorithm}[t]
\caption{PREFACE Prompt Adaptation Loop}
\label{alg:prompt_adaptation}
\begin{algorithmic}[1]
\Require Specification $\tau$  
Initial prompt $p_0 \gets \varphi(\tau)$ 
Max iterations $T_{\max}=7$ 
\For{$t = 0 \ldots T_{\max}-1$} 
    \State $c_t \gets \mathcal{M}(p_t)$ 
    \State $e_t \gets \mathrm{Verify}(c_t, \tau)$ 
    \If{$e_t = 0$} 
        \State \Return $c_t$ 
    \EndIf 
    \State $s_t \gets \psi(p_t, c_t, e_t)$ 
    \State Sample prompt edit: $\Delta p_t \sim  \pi_\theta(\,\cdot\,\mid s_t)$ 
    \State $p_{t+1} \gets p_t \oplus \Delta p_t$
\EndFor
\end{algorithmic}
\end{algorithm}

\subsection{MDP Formulation}
We formalize prompt refinement as an MDP \(\mathcal{G}=(\mathcal{S},\mathcal{A},P,R,\gamma)\):
\begin{itemize}
  \item \textbf{State} \(s_{t}=\psi(p_{t},c_{t},o_{t})\) encodes the current prompt, generated code, and verifier feedback. $(s_{t} \in \mathcal{S})$
  \item \textbf{Action} \(a_{t}\) is a discrete, token‑level modification \(\Delta p_{t}\) drawn from the SLM’s vocabulary. $(a_{t} \in \mathcal{A})$
  \item \textbf{Transition Dynamics} The distribution \(P(s_{t+1}\mid s_t,a_t)\) is implicitly defined by the combined generative and verification steps. Each state transition results from prompt modifications and their verification outcomes, implicitly defining the probabilistic transition model.
  \item \textbf{Reward} 
    \[
      R(s_{t},a_{t}) = 
      \begin{cases}
        +R_{\mathrm{succ}}, & e_{t+1}=0,\\
        -\alpha\,e_{t+1}-\beta, & e_{t+1}>0,
      \end{cases}
    \]
    with \(\alpha=0.2\), \(\beta=0.5\), and a large penalty for empty outputs.
  \item \(\gamma=0.99\) emphasizes long‑term verification success.
\end{itemize}

\subsection{Policy Optimization via PPO}
PREFACE employs Proximal Policy Optimization (PPO) to train the SLM agent. The actor maximizes the clipped surrogate objective
\[
L_{\mathrm{actor}}(\theta)
= -\mathbb{E}_{t}\bigl[\min\bigl(r_{t}(\theta)\,\hat{A}_{t},\,\mathrm{clip}(r_{t}(\theta),1-\epsilon,1+\epsilon)\,\hat{A}_{t}\bigr)\bigr],
\]
while the critic minimizes
\(\,L_{\mathrm{critic}}(\phi)=\mathbb{E}_t\bigl[(V_{\phi}(s_{t})-G_{t})^{2}\bigr]\).  We use Adam with gradient‑norm clipping for stable convergence. By restricting training to the lightweight SLM, PREFACE achieves formally verified code synthesis without fine‑tuning the base LLM, forming the core of Proof2Silicon.

\begin{figure*}[t]
    \centering
    \includegraphics[width=1\linewidth]{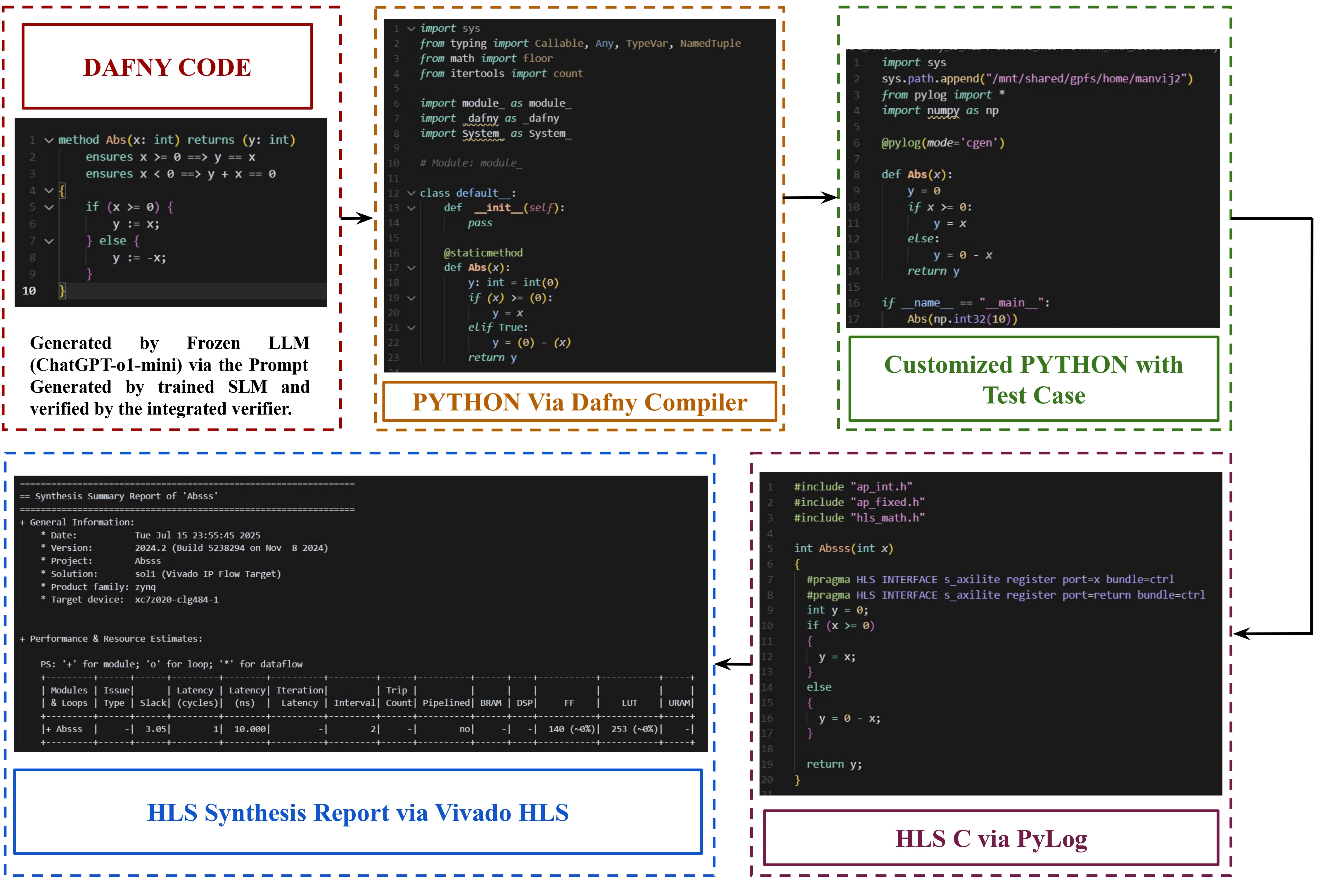}
    \caption{Proof2Silicon end‑to‑end workflow: an RL‑driven Small Language Model (SLM) iteratively refines meta‑prompts to a frozen LLM under guidance from the integrated Dafny verifier, producing formally verified Dafny code. Verified code is then compiled to Python, transformed into hardware‑aware C via the PyLog backend, and synthesized to RTL with Vivado HLS on a Zynq‑7000 SoC.}
    \label{fig:prooftosilicon_framework}
\end{figure*}

\section{The Complete Proof2Silicon Flow}

The Proof2Silicon method embeds the previously developed PREFACE \cite{10.1145/3716368.3735300} framework to facilitate hardware realization through the conversion of formally verified Dafny code to High-Level Synthesis (HLS) C code and further to RTL. Overall, Proof2Silicon encompasses a sophisticated, automated pipeline from Dafny code verification to FPGA hardware designs. Below, we detail each step of this advanced pipeline, emphasizing the intricate technicalities involved to ensure the generation of robust, verifiable, and synthesizable hardware code. The overview of the Proof2Silicon flow is shown in Figure \ref{fig:proposed_framework}

\subsection{Dafny to Python Translation}
Upon successful obtaining formally verified Dafny code through the PREFACE pipeline, the verified Dafny code undergoes a transformation into Python code using Dafny's native compiler with the command: \texttt{dafny build --target\:py verified.dfy}

This transformation is necessary as direct translation from Dafny to C incurs critical issues, particularly concerning integer overflow and limited support for various mathematical constructs required for accurate hardware synthesis.

\subsection{Removing Dafny-Specific Dependencies}
The generated Python code contains significant dependencies on Dafny's internal libraries, specifically designed for mathematical operations and array manipulations that adhere strictly to Dafny semantics. These dependencies introduce complexity that complicates HLS translation due to the incompatible semantics and the inclusion of empty class wrappers generated by Dafny's Python backend.

To address this challenge, we implemented an automated script that systematically removes Dafny-specific library imports and eliminates redundant class structures and wrappers. The script replaces Dafny-based mathematical operations with equivalent Python-native implementations, leveraging the NumPy library for numerical operations. This meticulous removal ensures syntactic simplicity and semantic clarity required for downstream synthesis.

\subsection{Customization with PyLog Decorators}
The sanitized Python code undergoes further refinement through integration with PyLog - an algorithm-centric Python-based FPGA programming and synthesis tool \cite{9591456}. PyLog facilitates the translation of Python to high-quality synthesizable HLS C through an intermediate representation (IR), extensive compiler optimizations, and automated design space exploration \cite{5226333}. Specifically, PyLog employs a pragma-based optimization mechanism that enables loop transformations, memory optimizations, and resource management to ensure efficient FPGA hardware synthesis.

We augment the Python scripts with PyLog decorators that explicitly annotate functions intended for hardware acceleration, allowing precise control over synthesis directives, including loop pipelining, unrolling, and array partitioning. These decorators guide the PyLog compiler toward generating optimized HLS code aligned closely with FPGA hardware execution models.

\paragraph{Ensuring Numerical and Data Type Compatibility}
Incorporating NumPy library calls, all numerical data types and operations in the Python code are explicitly converted to NumPy types (e.g., np.int32, np.float32). This rigorous type annotation is vital for numerical stability, predictable hardware performance, and compatibility with the fixed-point arithmetic typically utilized in FPGA designs and for ease of conversion via the PyLog tool.

\paragraph{Manual Test Case Development}
Effective synthesis via PyLog necessitates the explicit presence of test cases to validate synthesized designs and to ensure correctness in hardware execution. Given the limited availability of automated test generation tools suitable for this specific context, we manually crafted test cases for each DafnyBench \cite{loughridge2024dafnybenchbenchmarkformalsoftware} benchmark used in our evaluations. These cases thoroughly exercise the synthesized designs, validating both functional correctness and FPGA resource optimization.

\subsection{Python to Synthesizable C via PyLog}
After preparation with PyLog decorators and numerical optimizations, the refined Python code is translated into synthesizable C (HLS C) using PyLog’s compilation engine. PyLog optimizes the generated HLS C through aggressive pragma insertion, loop restructuring, and memory customization strategies, complementing scalability-oriented frameworks such as ScaleHLS \cite{9773203}. This comprehensive optimization ensures that the resulting HLS C efficiently maps onto FPGA hardware resources, enhancing throughput and minimizing latency.

\subsection{FPGA Hardware Synthesis and Verification}
The final stage uses Xilinx Vivado HLS to synthesize the generated C kernel into RTL, producing detailed reports on timing (clock‑period estimates, initiation interval) and resource utilization. Although we do not perform functional co‑simulation, this step verifies that the design is structurally FPGA‑compatible and meets target performance and area constraints.

In this way, Proof2Silicon seamlessly bridges formally‑verified Dafny specifications with practical hardware synthesis—automatically generating synthesizable RTL from natural language descriptions while providing confidence in timing and resource metrics.

Figure~\ref{fig:prooftosilicon_framework} illustrates the end‑to‑end Proof2Silicon workflow. First, the embedded PREFACE framework uses a trained Small Language Model (SLM) to generate meta‑prompts that guide a frozen LLM in producing candidate Dafny programs. Each program is immediately checked by an integrated Dafny verifier, and only those that pass formal verification proceed. Verified Dafny code is then compiled to Python using Dafny’s compiler, yielding code that still contains Dafny‑specific constructs. This Python code is then reformatted via our automation script to be converted to a PyLog-friendly script. It is then fed into our PyLog transformation pipeline, which rewrites it into HLS C kernels. Finally, these kernels are synthesized with Vivado HLS (10 ns target clock) on the xc7z020clg484‑1 device, producing RTL IP cores along with detailed timing and resource utilization reports. By coupling RL‑driven prompt optimization with formal verification and automated high‑level synthesis, Proof2Silicon delivers correctness‑by‑construction hardware directly from high‑level algorithmic descriptions.

\section{Results and Evaluation}

In this section, we provide an end‑to‑end evaluation of the Proof2Silicon workflow, beginning with PREFACE’s formally verified Dafny code synthesis results and concluding with the synthesis readiness and performance metrics of the generated HLS C codes. We evaluated the performance of several state-of-the-art language models in a representative subset of 100 tasks sampled from DafnyBench~\cite{loughridge2024dafnybenchbenchmarkformalsoftware}. The models tested include ChatGPT-4o \cite{openai2024gpt4technicalreport}, ChatGPT-o1 \cite{zhong2024evaluationopenaio1opportunities}, Qwen2.5-Coder-14B \cite{hui2024qwen25codertechnicalreport}, Qwen2.5-7B \cite{qwen2025qwen25technicalreport} and Gemini-2-Flash.


\subsection{Embedded PREFACE results}

\begin{table*}[ht]
\centering
\renewcommand{\arraystretch}{1.1}
\caption{Verification success rates on a 100‐task benchmark for various LLMs using prompts through PREFACE pipeline from both untrained and ~100-hr-trained small language model, comparing single‐shot (W/O Feedback) versus iterative error‐feedback (With Feedback) modes. We employ the "One-line and Detailed Description" prompting mechanism; values are color‐coded green for improvements and red for degradations relative to their  respective baseline values.}

\label{tab:proposed_results}
\begin{tabular}{|l|c|c|c|c|c|c|}
\hline
\textbf{Model} &
\multicolumn{2}{c|}{\textbf{Without SLM (baseline)}} &
\multicolumn{2}{c|}{\textbf{Without Trained SLM}} &
\multicolumn{2}{c|}{\textbf{With Trained SLM}} \\
\cline{2-7}
& \textbf{W/O Feedback} & \textbf{With Feedback} 
& \textbf{W/O Feedback} & \textbf{With Feedback} 
& \textbf{W/O Feedback} & \textbf{With Feedback} \\
    \midrule
    ChatGPT-4o              & 25\% & 36\% & 10\% (\textcolor{red}{-15\%}) & 44\% (\textcolor{forestgreen}{+8\%}) & 23\%  (\textcolor{red}{-2\%})& 50\% (\textcolor{forestgreen}{+14\%})\\
    \hline
    ChatGPT-o1-mini             & 23\% & 35\% & 8\% (\textcolor{red}{-14\%})& 42\% (\textcolor{forestgreen}{+7\%})& 23\% (\textcolor{forestgreen}{±0\%})& 52\% (\textcolor{forestgreen}{+17\%})\\
    \hline
    Qwen2.5-Coder-14B                  & 15\% & 21\% & 2\% (\textcolor{red}{-13\%})& 12\% (\textcolor{red}{-9\%})& 16\% (\textcolor{forestgreen}{+1\%})& 31\% (\textcolor{forestgreen}{+10\%})\\
    \hline
    Qwen2.5-7B               & 3\% & 7\% & 0\% (\textcolor{red}{-3\%})& 4\% (\textcolor{red}{-3\%})& 3\% (\textcolor{forestgreen}{±0\%}) & 11\% (\textcolor{forestgreen}{+4\%})\\
    \hline
    Gemini-2-Flash             &20\% & 33\% & 3\% (\textcolor{red}{-17\%})&  32\% (\textcolor{red}{-1\%})& 29\% (\textcolor{forestgreen}{+9\%}) & 55\% (\textcolor{forestgreen}{+21\%})\\
    \bottomrule
\end{tabular}
 
\end{table*}



\begin{table*}[ht]
\centering
\caption{Proof2Silicon Hardware Synthesis Results: Compilation and synthesis statistics for verified Dafny programs across LLMs and verifier feedback settings.}
\label{tab:proof2silicon_results}
\begin{tabular}{|l|c|c|c|c|c|c|}
\hline
\textbf{Model} & \textbf{Feedback} & \textbf{Verified Count} & \textbf{Compiled to HLS} & \textbf{HLS Synthesized} & \textbf{Avg. Elpased Time} & \textbf{Avg. Peak Memory} \\
\hline
ChatGPT-4o & No & 23 & 17 & 12 (52.1\%) & 32.98 sec & 705.41 MB\\
ChatGPT-4o & Yes & 50 & 29 & 25 (50\%) & 32.96 sec & 678.51 MB\\
ChatGPT-o1-mini & No & 23 & 17 & 12 (52.1\%) & 32.83 sec & 677.47 MB\\
ChatGPT-o1-mini & Yes & 52 & 35 & 29 (55.8\%) & 32.84 sec & 682.46 MB\\
Qwen2.5-Coder-14B & No & 16 & 12 & 10 (62.5\%) & 33.98 sec & 678.56 MB\\
Qwen2.5-Coder-14B & Yes & 31 & 23 & 17 (54.8\%)  & 33.47 sec & 681.97 MB\\
Qwen2.5-7B & No & 3 & 3 & 2 (66.67\%) & 33.51 sec & 685.46 MB\\
Qwen2.5-7B & Yes & 11 & 7 & 6 (54.54) & 32.76 sec & 677.746 MB\\
Gemini-2-Flash & No & 29 & 24 & 21 (72.4\%) & 33.88 sec & 679.98 MB \\
    Gemini-2-Flash & Yes & 55 & 44 & 38 (69.1\%) & 34.21 sec & 683.36 MB \\
\hline
\end{tabular}
\end{table*}


\begin{table*}[ht]
\centering
\caption{Performance and resource results for selected benchmarks on the Xilinx Zynq-7000 SoC.}
\resizebox{\textwidth}{!}{%
\begin{tabular}{| l| c| c| c| c| c| c| c| c| c| c| c| c| c| c|}
\hline
&  \multicolumn{12}{c|}{\textbf{Benchmark Tasks}}\\
\cline{2-13}
\textbf{Model (Feedback)} &  \multicolumn{4}{c|}{\textbf{Cube}} & \multicolumn{4}{c|}{\textbf{TriangleNumber}} & \multicolumn{4}{c|}{\textbf{TriangularPrismVolume}} \\
\cline{2-5} \cline{6-9} \cline{10-13} 
& \textbf{Latency} & \textbf{LUTs} & \textbf{DSPs} & \textbf{FFs}& \textbf{Latency} & \textbf{LUTs} & \textbf{DSPs} & \textbf{FFs}& \textbf{Latency} & \textbf{LUTs} & \textbf{DSPs} & \textbf{FFs} \\
\hline
ChatGPT-4o (No)           & 60 ns& 685  & 6  & 789 & 50 ns & 511  &3 & 436 & 60 ns & 456  &6  & 807 \\
ChatGPT-4o (Yes)          & 60 ns & 685 & 6  & 789   & 50 ns& 511 & 3  & 436  & 60 ns  & 546   & 6  & 807 \\
ChatGPT-o1-mini (No)      & 70 ns & 305 & 6 & 538   & 50 ns  & 576 & 3  & 436 & 60 ns & 546  & 6 & 807    \\
ChatGPT-o1-mini (Yes)     & 60 ns & 685  & 6 & 789  & 40 ns & 462  & 3  & 437 & 250 ns & 2127  & 11 & 908   \\
Gemini-2-Flash (No)       & 70 ns  & 798 & 6 & 880   & 50 ns  & 576   & 3  & 436 & 60 ns  & 546   & 6  &807   \\
Gemini-2-Flash (Yes)      & 80 ns & 381  & 6  & 572   & 40 ns & 462  & 3 & 437 & 290 ns & 1051  & 11  & 838 \\
Qwen2.5-Coder-14B (No)    & 70 ns  & 798  & 6 & 880  & 50 ns & 576  & 3   & 436 & 60 ns & 546   & 6  &807  \\
Qwen2.5-Coder-14B (Yes)   & 60 ns & 685  & 6 & 789  & 50 ns & 576 & 3 & 436 &  290 ns & 1051  & 11  & 838  \\
Qwen2.5-7B (No)           & 80 ns & 381 & 6 & 572    & 50 ns & 576 & 3  & 436 & 60 ns & 546   & 6  &807  \\
Qwen2.5-7B (Yes)          & 60 ns & 685& 6 & 789  & 40 ns & 462  & 3 & 437    & 290 ns & 1051  & 11  & 838 \\
\hline
\end{tabular}%
}
\label{tab:latency_values_transposed}
\end{table*}



\begin{figure}[ht]
  \centering
  \begin{subfigure}[b]{0.4\textwidth}
    \includegraphics[width=\textwidth]{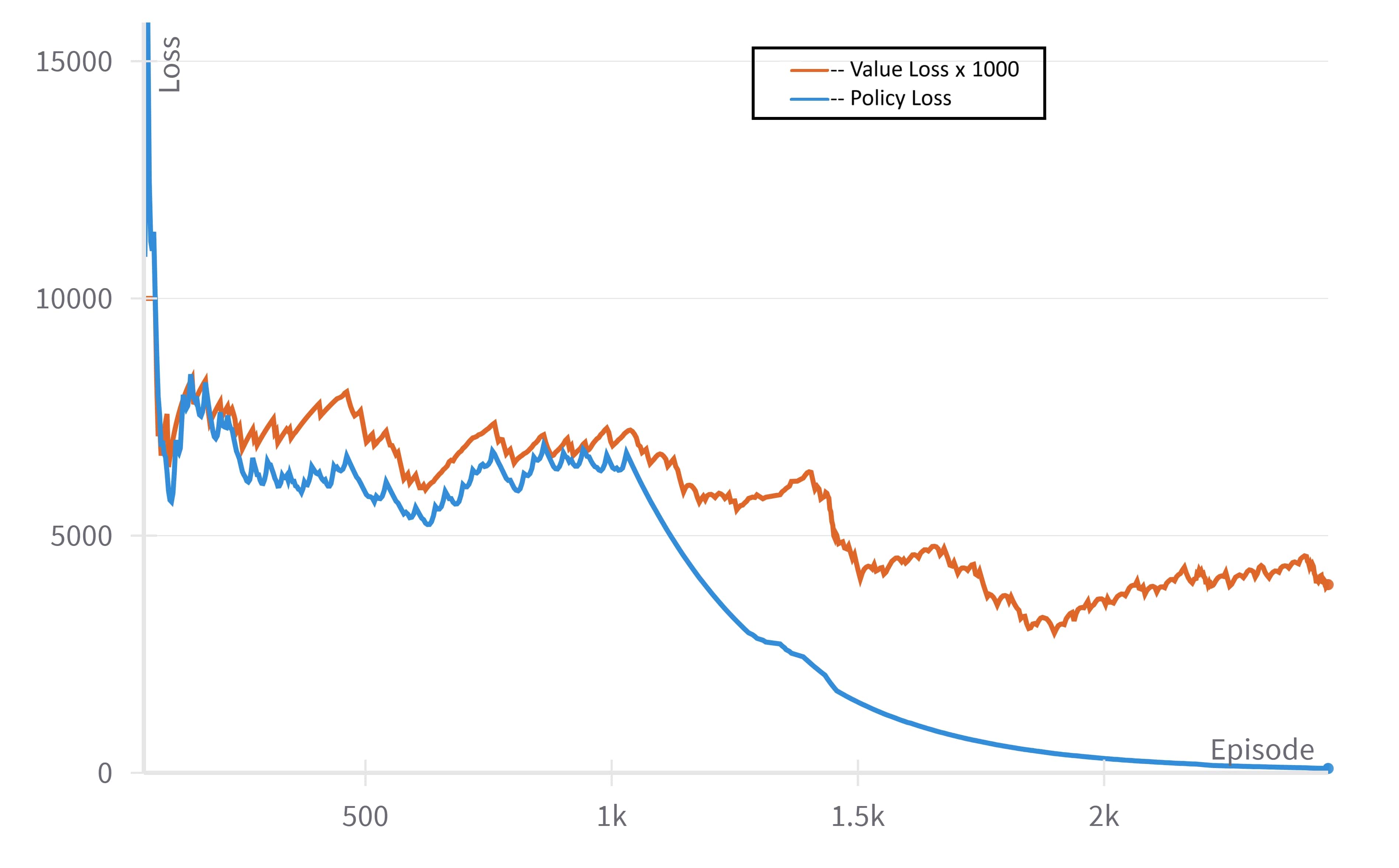}
    \caption{Value Loss $\times$ 1000 and Policy Loss vs.\ episode}
    \label{fig:total_loss}
  \end{subfigure}
  \hfill
  \begin{subfigure}[b]{0.4\textwidth}
    \includegraphics[width=\textwidth]{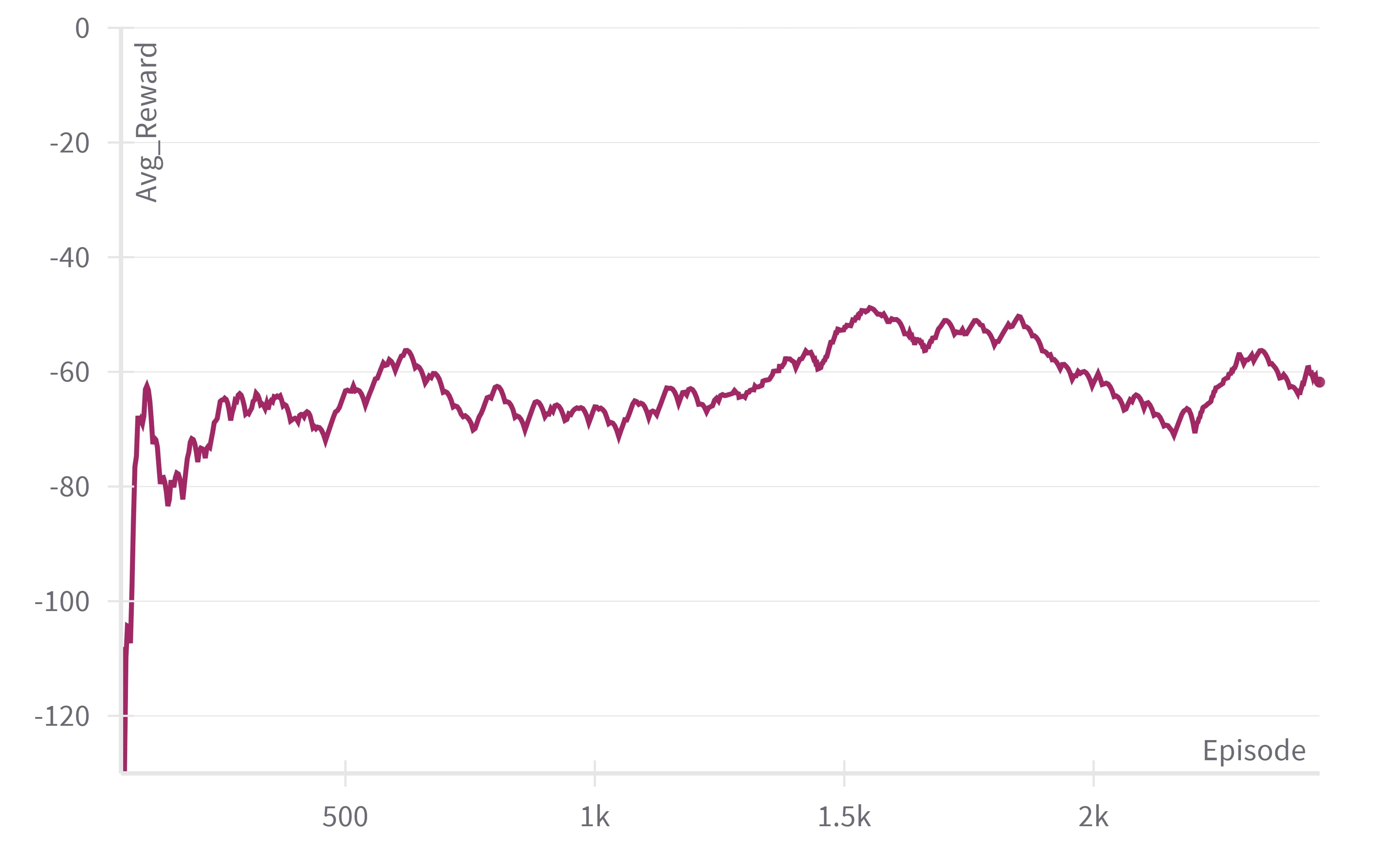}
    \caption{Episode reward vs.\ episode }
    \label{fig:reward}
  \end{subfigure}
  \caption{Training curves for the PREFACE pipeline over more than 2000 episodes: (a) Value Loss $ \times$ 1000 and Policy Loss, we scale up the Value loss for a better visualization; (b) Episode Reward, illustrates the pipeline's verification performance improving over time.}
  \label{fig:loss_reward}
\end{figure}

PREFACE was trained for approximately 100 h on two NVIDIA H100 GPUs, during which shaped rewards, policy loss, and value loss were recorded over 2000+ episodes (Figure~\ref{fig:loss_reward}). Both policy and value losses steadily declined, while episode rewards rose despite early‐stage penalties, indicating effective convergence and improved prompt optimization.

On a 100‐task DafnyBench subset (Table~\ref{tab:proposed_results}), static prompting baselines were outperformed by SLM‑tuned prompting even without feedback, achieving 23\% for both ChatGPT‑4o and ChatGPT‑o1, 16\% for Qwen2.5‑Coder‑14B, 3\% for Qwen2.5‑7B, and 29\% for Gemini‑2‑Flash. Enabling the full verifier feedback loop further boosted these rates to 50\%, 52\%, 31\%, 11\%, and 55\%, respectively, demonstrating PREFACE’s consistent, model‑agnostic enhancements in formal verification success.

\subsection{HLS Synthesis and Performance Results}

Extending beyond formal verification, we evaluated the Proof2Silicon pipeline’s ability to translate verified Dafny code into synthesizable HLS C and validate synthesis on FPGA platforms. We analyzed hardware compilation success, resource utilization, and performance metrics for each LLM and feedback setting. For our experiments, we simulate the synthesis of the HLS code on Zynq-7000 All Programmable SoC family (device: xc7z020clg484-1). The xc7z020 is a mid-range Zynq FPGA widely used in embedded applications requiring tight integration between programmable logic and embedded processors. It is well-suited for prototyping DSP, image processing, or control algorithms. For our experiment, the TCL scripts define a 10 ns clock period.

The results are presented in Table \ref{tab:proof2silicon_results}, demonstrating the effectiveness of the Proof2Silicon framework in translating verified Dafny programs into synthesizable hardware. The first stage reports the number of Dafny codes that successfully pass formal verification, establishing functional correctness. These verified programs are then translated to Python and passed through PyLog to produce HLS-compatible C, a stage we denote as Compiled to HLS. 

At this stage, many programs are excluded because PyLog does not support certain Dafny-derived constructs, most notably recursion, dynamic memory usage, or while-loops, all of which violate the requirement of acyclic and static control flow for hardware synthesis. Additional incompatibilities arise from Dafny-specific data structures and strict typing conventions, which, even after automated sanitization, may still produce kernels that cannot be transformed into valid HLS C. Among the codes that do compile, not all complete RTL synthesis in Vivado HLS, which imposes further restrictions such as prohibiting recursive function calls, requiring consistent function signatures, and rejecting kernels with subtle type mismatches (e.g., functions declared as void despite computing integer outputs). 

These issues account for the observed discrepancy between Compiled to HLS and HLS Synthesized, with approximately 30\% of otherwise verified programs excluded during this stage. For the designs that do pass through to full synthesis, we report three metrics: the total number of programs that successfully parse and elaborate in Vivado HLS, the subset that complete RTL synthesis, and the average per-kernel runtime and memory footprint. All experiments were conducted using unmodified TCL scripts with a fixed 10 ns target clock, and detailed resource and timing reports were collected from the Vivado HLS reporting interface. Together, these results highlight both the promise of Proof2Silicon in establishing a correctness-by-construction synthesis flow and the current limitations that must be addressed to improve alignment between verification and hardware realization.

As shown in Table \ref{tab:proof2silicon_results}, Proof2Silicon achieves robust synthesis readiness across diverse LLMs and feedback modes. Without verifier feedback, the pipeline synthesizes 52.1\% of ChatGPT‑o1-mini’s verified codes and up to 72.4\% for Gemini‑2‑Flash, with mean synthesis times of 32.8–33.9 s and peak memory under 705 MB. Enabling the full verifier feedback loop further improves synthesis rates—ChatGPT‑o1-mini rises to 55.8\%, while synthesis latencies remain stable (32.8–34.2 s) and memory usage exhibits minimal variance (678–683 MB). These results demonstrate that embedding PREFACE’s prompt‑optimization core within a hardware synthesis pipeline yields high end‑to‑end success: the majority of formally verified algorithms not only translate cleanly to HLS C but also compile and synthesize into FPGA‑ready RTL with predictable performance and resource consumption. Such consistency underlines Proof2Silicon’s potential for automated, correctness‑driven hardware generation in safety‑critical and high‑throughput applications.

It is important to note that the current results represent initial demonstrations of the Proof2Silicon flow rather than a fully optimized end-to-end solution. While the framework shows promising potential by achieving a high hardware synthesis success, several synthesis failures highlight the need for further refinement of the pipeline. In future work, we plan to enhance compatibility between Dafny translation, PyLog optimizations, and Vivado HLS, ensuring that the number of verified programs successfully synthesized into RTL aligns more closely with the number of verified inputs.

Table~\ref{tab:latency_values_transposed} summarizes the performance and resource results for selected benchmark tasks across different LLMs and verifier feedback settings, demonstrating that the successfully synthesized designs are executable on FPGA hardware. While our current focus is on correctness and verifiability, these preliminary numbers also confirm that the generated HLS C codes can be mapped onto the FPGA fabric with consistent resource utilization. Future work will extend beyond correctness to explicitly optimize synthesis performance metrics such as latency and resource utilization, enabling the generation of hardware-aware code that balances verification guarantees with practical throughput and efficiency in FPGA implementations.

\subsection{Implementation Details}
All experiments were conducted on a dedicated workstation equipped with two NVIDIA H100  GPUs, running Ubuntu 20.04 and Python 3.10. Evaluation was done on the DafnyBench suite- 100 test tasks covering algorithms such as sorting, arithmetic operations, and list manipulations.

\subsection{Future Work}
A key direction for future work is to enhance the compatibility between Dafny verification, PyLog translation, and Vivado HLS synthesis. We plan to refine the translation flow to proactively avoid unsupported constructs such as recursion, dynamic loops, and other control-flow patterns that currently hinder hardware generation, thereby reducing synthesis failures and narrowing the gap between the number of verified programs and those successfully synthesized. Furthermore, we aim to extend the framework beyond correctness by integrating hardware-aware optimizations directly into the PREFACE agent, enabling it to generate Dafny code that is both verification-friendly and synthesis-ready. Incorporating synthesis performance metrics such as latency, resource utilization, and throughput into the optimization objectives will allow Proof2Silicon to balance formal verification guarantees with practical hardware efficiency, ultimately enabling the generation of high-quality FPGA implementations.

\section{Conclusion}

We present Proof2Silicon, a comprehensive pipeline that extends PREFACE’s reinforcement learning–based prompt repair framework toward hardware synthesis. By coupling formally verified Dafny code generation with automated translation to synthesizable HLS C via PyLog, and validating synthesis with Vivado HLS, Proof2Silicon bridges formal methods and hardware design automation.

Our experiments show that PREFACE’s prompt refinement significantly increases verified Dafny code generation, while a meaningful subset of this verified code successfully compiles and synthesizes into hardware accelerators. We identify current limitations due to unsupported recursion and control flow structures in HLS, and highlight ongoing efforts to enhance pipeline robustness.


Proof2Silicon thus establishes a scalable, model-agnostic, and formally verified approach to trustworthy AI-assisted hardware synthesis, charting a promising path forward for next-generation EDA tools and AI-driven design automation.

\section*{Acknowledgement}
This work is supported by the AMD Center of Excellence grant at UIUC and Semiconductor Research Corporation (SRC) 2023-CT-3175 grant.


\bibliography{conference_101719}

\end{document}